\PassOptionsToPackage{hyphens}{url}
\documentclass{article}
\usepackage{spconf,amsmath,amssymb,graphicx}
\usepackage{enumitem}
\setlist{nosep, leftmargin=14pt}
\usepackage{booktabs}
\usepackage{url}
\usepackage{xcolor}
\usepackage{mwe} % dummy images for placeholders -- remove when real figures are in
\usepackage{float}
\usepackage[hyphens]{url}   % before hyperref
\usepackage{hyperref}
\hypersetup{breaklinks=true}

\def\L{{\cal L}}
\def\z{{\mathbf z}}
\def\c{{\mathbf c}}

\title{LungMend: A Healthy-Only Latent Diffusion Shape Prior \\ for Lung Mask Correction in severe Tuberculosis CT}

\name{Rezkellah Noureddine Khiati$^{1}$, Pierre-Yves Brillet$^{2}$, and Catalin Fetita$^{1}$}
\address{$^{1}$SAMOVAR, Télécom SudParis, Institut Polytechnique de Paris, Palaiseau, France \\
        $^{2}$Avicenne Hospital, AP-HP, Bobigny, France}

\begin{document}
\maketitle

% =========================================================================
\begin{abstract}
Lung segmentation models such as R231 are widely used, but they fail on severe tuberculosis (TB) CT: cavities and consolidations look like non-lung tissue and are left out of the mask. We present LungMend, a post-processing model that corrects such masks using only healthy anatomy. A latent diffusion model, trained on healthy lungs with synthetically corrupted masks, learns to recover the complete lung from the faulty mask and the bone anatomy, without any pathological data or CT intensities. LungMend recovers cavities and consolidations that R231 misses and improves accuracy on expert-annotated slices. Code: \href{https://github.com/noureddinekhiati/LungMend}{noureddinekhiati/LungMend}
\end{abstract}

\begin{keywords}
Lung segmentation, latent diffusion, shape prior
\end{keywords}

% =========================================================================
\section{Introduction}
\label{sec:intro}

Lung segmentation is the first step of most quantitative CT pipelines. The most widely used model is the open-source U-Net R231~\cite{hofmanninger2020lungmask}, the de facto standard in 3D Slicer. It was trained on a large and diverse set of 231 CT scans that includes consolidation and several other pathological patterns. Yet it still fails on TB, where large cavities and extensive consolidations have soft-tissue density and are excluded from the mask. We observe three recurrent failure modes: (i) loss of large contiguous regions, (ii) erosion along the chest wall where disease is subpleural, and (iii) slice-to-slice inconsistency, since R231 works in 2D and its 3D masks are fragmented along the craniocaudal axis.

\textbf{Related work.} Dedicated pathological lung segmentation combines intensity, texture and shape cues~\cite{sluimer2005pathological,mansoor2014generic,harrison2017pathological}, but it needs pathological annotations, which are costly and never exhaustive. Shape-based post-processing, from statistical shape models~\cite{heimann2009ssm} to autoencoder priors~\cite{oktay2018acnn,larrazabal2020postdae,painchaud2020cardiac}, projects a mask onto a single plausible shape and struggles when large regions are missing. Diffusion models~\cite{ho2020ddpm,song2021ddim} have been used to refine segmentations~\cite{wolleb2022ensembles,wu2023medsegdiff,wang2023segrefiner}, but they condition on image intensities, which are what misleads the segmenter in diseased lungs. Latent diffusion~\cite{rombach2022ldm} makes 3D medical generation tractable~\cite{pinaya2022brainldm,khader2023ddpm3d}, and shape diffusion enables 3D completion in vision~\cite{park2019deepsdf,zhou2021pvd,cheng2023sdfusion,zhang2023shape2vecset}. To our knowledge, no prior work uses a generative shape prior learned from healthy anatomy to correct pathological lung masks.

\textbf{Contributions.}
\begin{itemize}
  \item \textbf{New formulation.} We cast lung mask correction as shape completion with a prior learned from healthy anatomy only, so no pathological annotations are needed and the correction does not depend on the image appearance that causes R231 to fail.
  \item \textbf{Training without pathology.} We introduce a corruption model that reproduces the failure modes of texture-based segmenters, which makes it possible to train a correction model on healthy lungs alone.
  \item \textbf{Disease-independent anchoring.} We show that bone anatomy gives a reliable cue for completing large missing regions, since the rib cage is not affected by parenchymal disease.
  \item \textbf{Validation on real TB.} We evaluate on expert-annotated TB cases and show improvements over R231 and shape-prior baselines, including 3D consistency.
\end{itemize}
% =========================================================================
\begin{figure*}[t]
  \centering
  \includegraphics[width=\textwidth]
  {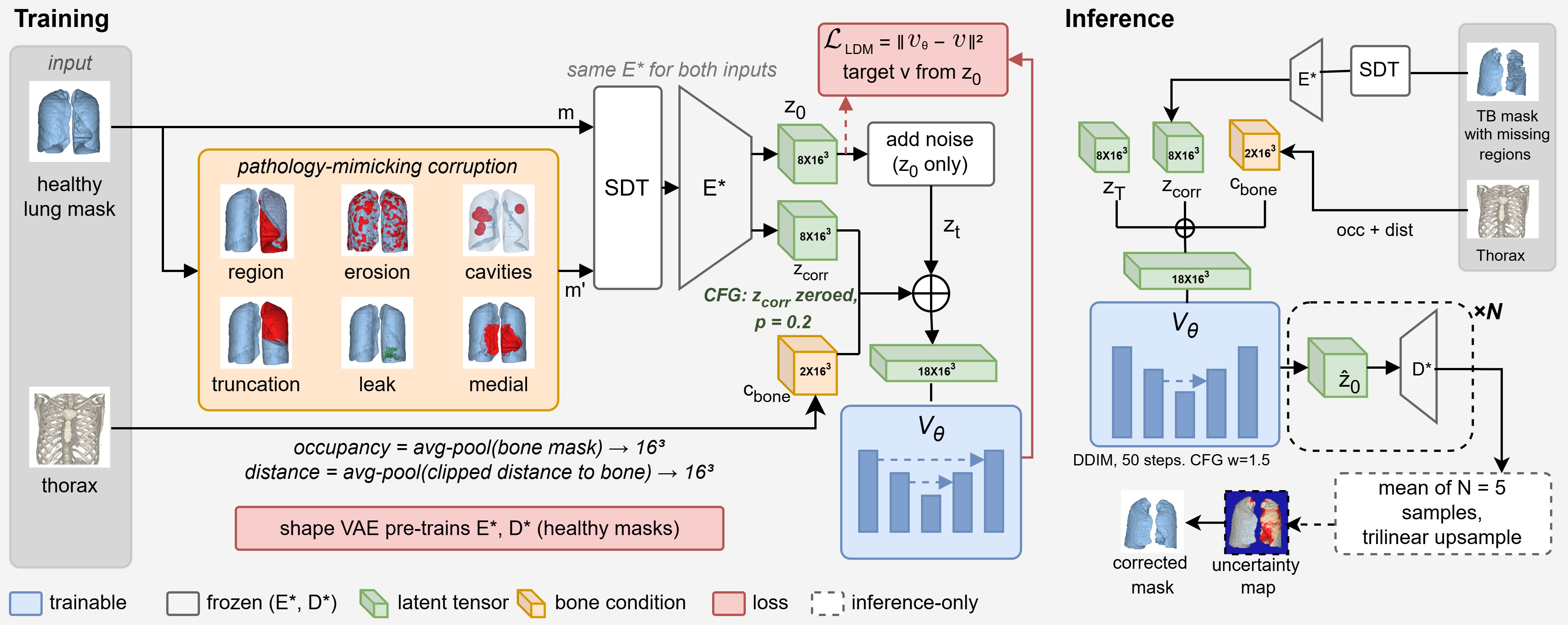}
  
  \caption{Overview of LungMend.
  \label{fig:pipeline}
  \textbf{Training (Stage 2):} the SDT of a healthy lung mask is encoded by the frozen VAE encoder $E^*$ (pre-trained in Stage~1) into $\z_0$. A pathology-mimicking corruption of the mask gives $\z_{\mathrm{corr}}$, and bone occupancy and distance give $\c_{\mathrm{bone}}$. The U-Net V$_{\theta}$ takes the noisy latent $\z_t$ and $[\z_{\mathrm{corr}},\c_{\mathrm{bone}}]$ (18 channels, $16^3$) and predicts $v_\theta$. \textbf{Inference:} $N$ samples are decoded by $D^*$ and averaged into the corrected mask and an uncertainty map. CFG denotes: classifier-free guidance. }
\end{figure*}

\section{Method}
\label{sec:method}

Given an incomplete lung mask $m'$ (e.g.\ the R231 output) and a bone mask $b$, LungMend estimates the complete lung mask $\hat m$. A model that has only seen healthy lungs knows what a lung \emph{should} look like, so it can fill the regions where disease made the mask deviate from that shape. To do this, the lung is represented as a smooth shape rather than as image intensities, which are what misleads the segmenter, and the missing region is anchored by the bone anatomy, which disease does not alter. LungMend is trained in two stages, both on healthy lungs only (Fig.~\ref{fig:pipeline}): Stage~1 learns a compact space of healthy shapes, and Stage~2 learns to recover a complete lung from a corrupted one. No pathological scan, mask or CT intensity is used.

\textbf{Data and shape representation.} Both stages use the same 57 healthy post-COVID scans from COVALUNG (COVALUNG dataset, Caen University Hospital, France). Each lung is cropped (bounding box $+15$\,mm), cleaned (two largest components, hole filling) and resampled to $128^3$; the bone mask gets the same crop. Instead of binary masks, we use the signed distance transform (SDT),
\begin{equation*}
  \mathrm{SDT}(m) = \mathrm{clip}\big(d(m)-d(\bar m),\,-C,\,C\big)/C,
\end{equation*}
with $d(\cdot)$ the Euclidean distance transform, $\bar m$ the complement of $m$ and $C{=}20$\,mm. We write $s=\mathrm{SDT}(m)$ for a complete mask and $s'=\mathrm{SDT}(m')$ for an incomplete one. Each voxel stores its signed distance to the lung surface (positive inside), and a mask is recovered as $m=[s>0]$. The SDT is smooth, so it can be compressed and denoised without the isolated-voxel artifacts of binary masks.

\subsection{Stage 1: shape autoencoder}
A 3D VAE~\cite{kingma2014vae} encodes $s$ ($1{\times}128^3$) into a latent $\z$ ($8{\times}16^3$) and decodes it to $\hat s$ ($1{\times}128^3$). The encoder is fully convolutional, with three stride-2 convolutions (total stride 8), so the latent is a \emph{spatial} grid that keeps the layout of the lung, not a single global vector. The loss is
\begin{equation}
  \L_{\mathrm{VAE}} = \|\hat s - s\|_1 + \lambda\,\L_{\mathrm{Dice}}(\hat s, s) + \beta\,\mathrm{KL}.
  \label{eq:vae}
\end{equation}
The first term preserves the global distance field. $\L_{\mathrm{Dice}}$ is a soft Dice on sigmoid-relaxed SDTs~\cite{xue2020shapeaware} that sharpens the surface. KL is the Kullback--Leibler divergence between the encoder distribution $q(\z|s)$ and a standard Gaussian $\mathcal{N}(0,I)$, which regularizes the latent space. We use $\lambda{=}1$ and $\beta{=}10^{-6}$. The VAE is then frozen ($E^*$, $D^*$); having seen only healthy lungs, its latent space describes healthy shapes only.

\subsection{Stage 2: anatomy-conditioned latent diffusion}
\textbf{Pathology-mimicking corruption.} Without pathological ground truth, we create (faulty, true) mask pairs from healthy lungs. For each training lung $m$, a corrupted mask $m'$ is generated on the fly with the failure modes of texture-based segmenters: \emph{region drop} (a contiguous region removed from the apex, base or lateral wall), \emph{medial drop}, \emph{subpleural erosion}, \emph{cavities} (internal holes) and \emph{truncation} (a lung cut by an oblique plane). Occasionally a \emph{leak} dilates the mask, so the model also learns to remove voxels. Most corruptions are mild, so the model learns to keep correct regions unchanged, and a new corruption is drawn at each epoch.

\textbf{Bone conditioning.} When a large region is missing, the mask alone admits many completions. We add information that does not depend on the disease: ribs and chest wall are unaffected by parenchymal lesions and tightly bound the lung. We provide the bone occupancy $o$ and a distance-to-bone field $\delta$, both pooled to $16^3$, as $\c_{\mathrm{bone}}=[o,\delta]$ (2 channels). Thin ribs largely vanish after pooling, so the smooth $\delta$ carries the distance to the chest wall.

\textbf{Diffusion model.} The frozen encoder gives $\z_0=E^*(s)$ and $\z_{\mathrm{corr}}=E^*(s')$. We add noise to the clean latent, $\z_t=\alpha_t\z_0+\sigma_t\epsilon$, where $t$ is the diffusion time, $\epsilon\sim\mathcal{N}(0,I)$, and $\alpha_t$, $\sigma_t$ are the signal and noise scales of a cosine schedule~\cite{nichol2021improved}. A 3D U-Net~\cite{cicek20163dunet} V$_{\theta}$ receives $\z_t$ (8 channels) and the condition $\c=[\z_{\mathrm{corr}},\c_{\mathrm{bone}}]$ ($8{+}2$ channels), i.e.\ 18 input channels at $16^3$, and predicts the velocity~\cite{salimans2022progressive} $v_\theta$ (8 channels). It is trained to match the target velocity $v=\alpha_t\epsilon-\sigma_t\z_0$:
\begin{equation}
  \L_{\mathrm{LDM}} = \mathbb{E}_{t,\epsilon}\big\|v_\theta(\z_t,t,\c) - v\big\|^2.
  \label{eq:ldm}
\end{equation}
During training, $\z_{\mathrm{corr}}$ is zeroed with probability 0.2 (bone channels always kept), so the same network also learns to predict from the bone alone. This prevents the model from copying its input and enables classifier-free guidance~\cite{ho2022cfg}: at inference we combine the two predictions as $v_w = v_{\mathrm{bone}} + w\,(v_{\mathrm{full}} - v_{\mathrm{bone}})$, where $v_{\mathrm{full}}=v_\theta(\z_t,t,\c)$ uses the input mask and $v_{\mathrm{bone}}=v_\theta(\z_t,t,[\mathbf 0,\c_{\mathrm{bone}}])$ uses only the bone prior. The guidance weight $w$ sets the balance: $w{=}1$ gives the full prediction, and larger $w$ keeps the output closer to the input mask.

\textbf{Inference and uncertainty.} The input mask $m'$ obtained with R231 and the bone mask are cropped, and $m'$ is converted to $s'$ and encoded into $\z_{\mathrm{corr}}$. We draw $N$ latents with DDIM~\cite{song2021ddim} (50 steps, guided with $w$) and decode each with $D^*$ into an SDT $\hat s^{(n)}$. The \emph{mean} of the $N$ decoded SDTs, $\bar s=\frac{1}{N}\sum_n \hat s^{(n)}$, is upsampled trilinearly to the native grid and thresholded at zero, and the two largest components form the corrected mask $\hat m$. The \emph{standard deviation} of the same $N$ SDTs, $u(x)=\mathrm{std}_n\,\hat s^{(n)}(x)$, gives a voxel-wise uncertainty map: it is close to zero where the samples agree, for example where the input mask is kept, and high where they differ, i.e.\ where several completions are plausible.

% =========================================================================

\section{Experiments}
\label{sec:exp}

\subsection{Data}
\textbf{Training data.} The VAE (Stage~1) and the diffusion model (Stage~2) are trained on the same private COVALUNG cohort: 57 healthy post-COVID patients with no consolidation or other parenchymal lesion (90/10\% patient-level split). No pathological scan or mask is used at any point.

\textbf{Test data.} We use 20 TB cases from ImageCLEF TB \cite{kozlovski2021imageclef}, selected with the DeepPulmo annotations \cite{tan2024deeppulmotb} as the most severe in cavity burden, since large cavities and consolidations are where R231 fails most. Cavities cover 3.1--16.1\% of the lung volume (largest cavity 57--962\,mL, 4--36 cavities per case) and consolidation 2.0--21.9\%. Patient lists and details are provided in the code repository. The DeepPulmo lung masks are not accurate enough to serve as a reference (examples in the repository), so we use only its lesion labels. To obtain a reliable reference, a thoracic radiologist corrected the R231 mask on 200 axial slices, 10 per case. Finally, we assess LungMend qualitatively on TB CT scans of 18 patients from Avicenne Hospital, Bobigny, France.

\subsection{Evaluation}
We evaluate in two ways:
\begin{itemize}
  \item \emph{Expert slices:} Dice and average surface distance (ASD) on the 200 radiologist-corrected slices.
  \item \emph{Lesion recovery:} using the DeepPulmo cavity and consolidation labels, the fraction of each lesion included in the predicted lung mask (recall), over the 200 slices.
\end{itemize}
In addition, we assess LungMend qualitatively on the Avicenne cohort.

\subsection{Baselines and implementation}
\textbf{Baselines.} \emph{Segmentation models applied to the CT:} R231, a U-Net trained on diverse routine CT including pathological cases; TotalSegmentator~\cite{wasserthal2023totalsegmentator}, an nnU-Net-based model trained on 1204 CT examinations of clinical routine, with varied pathologies, scanners and sites (lung output: union of the lung lobes); and VISTA3D~\cite{he2025vista3d}, a MONAI segmentation foundation model with a SegResNet backbone, trained on 11{,}454 CT volumes with anatomical and lesion annotations. \emph{Mask correction:} R231 with morphological closing (10\,mm ball) and hole filling; Post-DAE~\cite{larrazabal2020postdae} trained on the same healthy masks and corruptions; and a deterministic 3D U-Net that takes the CT, the R231 mask and the bone mask and predicts the corrected mask, trained on the same healthy scans and corruptions. It has the same conditioning as LungMend plus the CT, but no diffusion step, which isolates the effect of the generative formulation.

\begin{table}[!t]
\centering
\caption{Results on the 200 expert slices of severe TB cases (mean; ASD and HD95 in mm). Best in bold.}
\label{tab:main}
\resizebox{\columnwidth}{!}{%
\begin{tabular}{l ccc|cc}
\toprule
 & \multicolumn{3}{c|}{Expert slices} & \multicolumn{2}{c}{Lesion recall} \\
\cmidrule(lr){2-4}\cmidrule(lr){5-6}
Method & Dice$\uparrow$ & ASD$\downarrow$ & HD95$\downarrow$ & Cav.$\uparrow$ & Cons.$\uparrow$ \\
\midrule
R231~\cite{hofmanninger2020lungmask} & 0.909 & 4.38  & 27.4 & 0.869 & 0.732 \\
R231 + closing                       & 0.929 & 3.79  & 22.9 & 0.934 & 0.868 \\
Post-DAE~\cite{larrazabal2020postdae}& 0.768 & 11.87 & 36.5 & 0.968 & 0.942 \\
Det.\ 3D U-Net                       & 0.896 & 5.04  & 26.1 & 0.557 & 0.731 \\
TotalSegmentator                     & 0.919 & 4.29  & 24.7 & 0.951 & 0.723 \\
VISTA3D                              & 0.847 & 7.61  & 38.8 & 0.726 & 0.364 \\
\textbf{LungMend (ours)}             & \textbf{0.986} & \textbf{0.57} & \textbf{5.5} & \textbf{0.993} & \textbf{0.981} \\
\bottomrule
\end{tabular}}
\end{table}

% ---------- Table 2 (single column) ----------
\begin{table}[t]
\centering
\caption{Ablations on the 200 expert slices (ASD in mm).
\textbf{A}: one training component removed at a time, with the model retrained using the same protocol as the full recipe; all rows use plain sampling (1 sample, $w{=}1$).
\textbf{B}: inference steps added one at a time to the final model.
R231 input: Dice 0.909.}
\label{tab:ablation}
\footnotesize
\setlength{\tabcolsep}{5pt}
\renewcommand{\arraystretch}{1.05}
\begin{tabular}{@{}lcc@{}}
\toprule
 & Dice$\uparrow$ & ASD$\downarrow$ \\
\midrule
\multicolumn{3}{@{}l}{\textit{A. Training components}} \\
Full recipe (reference)            & 0.964 & 2.34 \\
\cmidrule(lr){1-3}
Latent $\to$ voxel-space diffusion & 0.917 & 4.54 \\
SDT $\to$ binary mask$$    & 0.795 & 15.02 \\
No Dice term in VAE                & 0.935 & 3.68 \\
\cmidrule(lr){1-3}
No bone mask                       & 0.929 & 3.47 \\
No distance-to-bone                & 0.926 & 3.75 \\
No lung-channel dropout            & 0.939 & 3.32 \\
\cmidrule(lr){1-3}
No random-blob corruptions            & 0.923 & 4.02 \\
No medial corruption               & 0.938 & 3.16 \\
No leak corruption                 & 0.937 & 3.49 \\
\midrule
\multicolumn{3}{@{}l}{\textit{B. Inference (final model, cumulative)}} \\
Plain sampling (1 sample, $w{=}1$) & 0.964 & 2.34 \\
\quad + guidance $w{=}1.5$         & 0.967 & 2.15 \\
\quad + average of 5 samples       & 0.976 & 0.93 \\
\quad + smooth SDT upsampling (= LungMend) & \textbf{0.986} & \textbf{0.57} \\
\bottomrule
\end{tabular}
\end{table}

\begin{figure*}[!t]
\centering
  \centering
  \includegraphics[width=\textwidth]{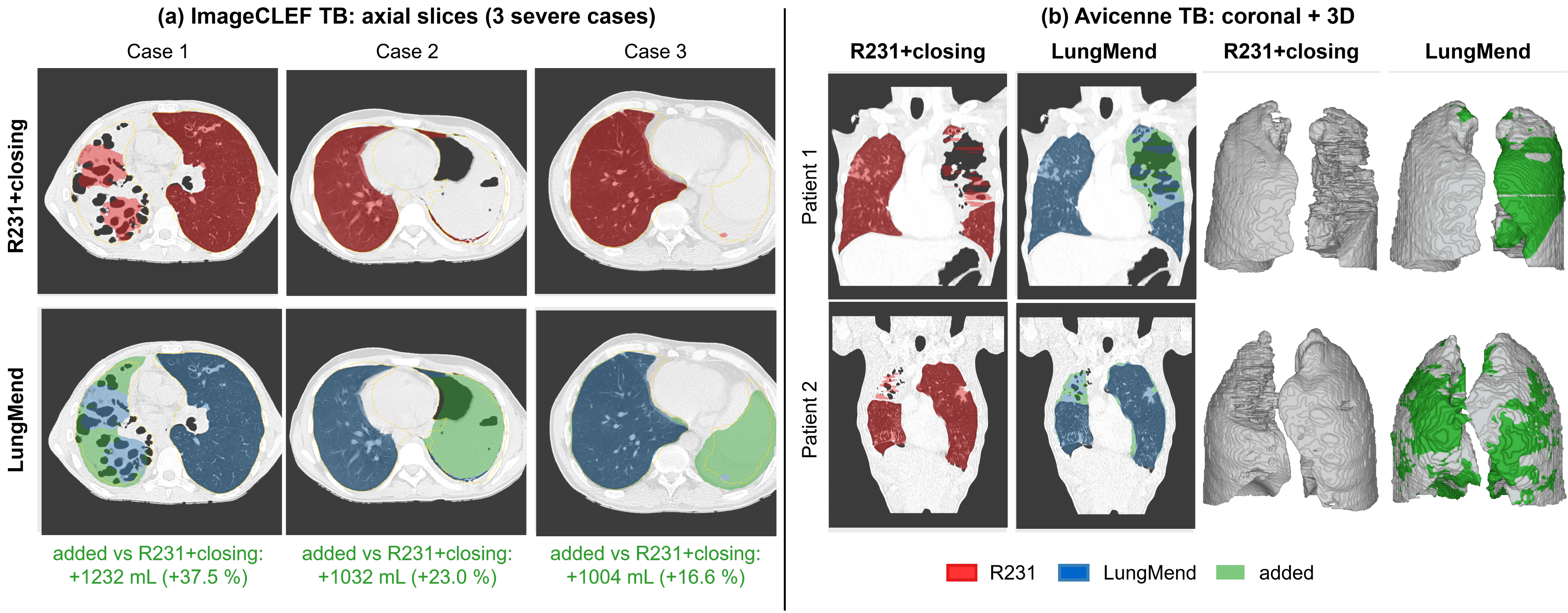}
  \caption{Qualitative results on severe TB. (a)~ImageCLEF TB axial slices: R231+closing (red, top) and LungMend (blue, bottom) masks overlaid on the CT, with the  annotation in yellow. The region added by LungMend is green; the volume added over R231 is given under each case. (b)~Avicenne TB: coronal slices and 3D renderings for R231 and LungMend. LungMend recovers cavities and consolidated lung that R231 excludes, and removes the stair-step fragmentation of the R231 3D surface.}
  \label{fig:qual}
\end{figure*}

\section{Results and discussion}
\label{sec:results}

\textbf{Comparison with baselines.} LungMend is the best method on every metric of Table~\ref{tab:main}, with the largest margin on HD95 (5.5\,mm vs.\ 22.9\,mm for the strongest baseline, R231 + closing): the baselines are accurate on most of the lung but leave large regions out, which dominates the worst-case distance. General segmentation models, even when trained on pathological CT, still decide from appearance, and more training data does not fix this (VISTA3D does worse than TotalSegmentator). Closing improves the surface but cannot recreate missing regions, and Post-DAE recovers lesions but over-segments, since it has no anatomical anchor. The deterministic U-Net has all the inputs of LungMend plus the CT, so it isolates the diffusion step: when several completions are plausible, a deterministic model returns a single estimate and loses the cavities, whereas diffusion samples plausible completions and averages them.

\textbf{Number of samples and inference time.} We tested several values of $N$ and chose $N{=}5$: with fewer samples the averaged map is sparse and unstable across runs, while beyond $5$ we observed no noticeable change. The cost grows linearly with $N$, but the samples are independent and can run in parallel on the GPU, so $N{=}5$ remains practical as a one-off post-processing step per scan.

\textbf{Ablation.} The ablation (Table~\ref{tab:ablation}, single runs) shows that the information given to the model matters more than architectural choices. Bone conditioning helps where the mask alone leaves a missing region ambiguous, and pathology-mimicking corruptions are the condition for transfer: a model trained on random blobs loses the most lesion recall, because a healthy-only prior is useful only if the simulated failures resemble the real ones. Latent diffusion on the SDT beats voxel-space diffusion and binary masks. The medial and leak corruptions and the conditioning dropout have a clear effect here. At inference, guidance and averaging reduce the variability between samples, and a single sample is already competitive.

\textbf{Qualitative results.} Fig.~\ref{fig:qual} compares LungMend with R231 + closing on cases selected by the volume LungMend adds. The added regions are the large cavities and consolidations on the diseased side; further examples on Avicenne scans are in Fig.~\ref{fig:qual}b and in the repository.

\textbf{Limitations.} A healthy prior may over-complete lungs with genuine volume loss (fibrotic retraction, destroyed lung), so the uncertainty map should be read together with the input mask. The gain depends on corruptions that resemble real failures, and the $128^3$ resolution limits precision near the hilum. Lesion recall depends on the DeepPulmo labels, and the expert slices cover only part of the lung.

% =========================================================================

% ---------- Fig. 2 ----------

\section{Conclusion}
\label{sec:conclusion}

We presented LungMend, a post-processing model that corrects lung masks in severe TB CT using only healthy anatomy. A latent diffusion prior trained on healthy lungs, together with a corruption model that reproduces the failure modes of texture-based segmenters and a bone-anchored condition, recovers cavities and consolidations that R231 excludes, without any pathological training data or CT intensities. Repeated sampling also provides an uncertainty map that shows which regions of the corrected mask were reconstructed rather than kept from the input.

% =========================================================================
% Everything below goes on page 5 (non-technical content only).
\clearpage

\section{Acknowledgments}
\label{sec:acknowledgments}
This work was carried out within a ANRS-MIE project No. ANRS00885-R-PR and 
ANR MLQ-CT project No. ANR-23-CE17-0046

\bibliographystyle{IEEEbib}
\bibliography{strings,refs}

\end{document}